\documentclass{article}

\usepackage{colt10e}
\usepackage{times}
\usepackage{amsfonts}
\usepackage{amsmath}
\usepackage[psamsfonts]{amssymb}
\usepackage{latexsym}
\usepackage{color}
\usepackage{graphics}
\usepackage{enumerate}
\usepackage{amstext}
\usepackage{url}
\usepackage{epsfig}

\DeclareMathOperator*{\E}{\rm E}

\DeclareMathOperator{\Tr}{Tr}

\newcommand{\mat}[1]{{\mathbf #1}}
\newcommand{\K}{\mat{K}}

\renewcommand{\P}{\mat{\Phi}}
\newcommand{\h}{\widehat}
\newcommand{\R}{\mathfrak{R}}

\newcommand{\w}{\mat{w}}

\newcommand{\1}{\mat{1}}

\newcommand{\Alpha}{{\boldsymbol \alpha}}

\newcommand{\Mu}{{\boldsymbol \mu}}

\newcommand{\Tau}{{\boldsymbol \tau}}

\newcommand{\set}[1]{\{#1\}}
\newcommand{\ignore}[1]{}

\title{New Generalization Bounds for Learning Kernels}

\author{Corinna Cortes\\
Google Research \\ 
New York\\
\texttt{\small corinna@google.com}
\And Mehryar Mohri\\
Courant Institute and \\
Google Research \\
\texttt{\small mohri@cims.nyu.edu}
\And Afshin
Rostamizadeh\\
Courant Institute\\
New York University\\
\texttt{\small rostami@cs.nyu.edu}}

\begin{document}
\maketitle

\begin{abstract}
  This paper presents several novel generalization bounds for the
  problem of learning kernels based on the analysis of the Rademacher
  complexity of the corresponding hypothesis sets. Our bound for
  learning kernels with a convex combination of $p$ base kernels has
  only a $\log p$ dependency on the number of kernels, $p$, which is
  considerably more favorable than the previous best bound given for
  the same problem. We also give a novel bound for learning with a
  linear combination of $p$ base kernels with an $L_2$ regularization
  whose dependency on $p$ is only in $p^{1/4}$.

\end{abstract}

\section{Introduction}

Kernel methods are widely used in statistical learning
\cite{kernel_book_1,kernel_book_2}. Positive definite symmetric (PDS)
kernels specify an inner product in an implicit Hilbert space where
large-margin methods are used for learning and estimation. They can be
combined with algorithms such as support vector machines (SVMs)
\cite{bgv,ccvv,vapnik98} or other kernel-based algorithms to form
powerful learning techniques.

But, the choice of the kernel, which is critical to the success of the
algorithm, is typically left to the user. Rather than requesting the
user to commit to a specific kernel, which may not be optimal for the
task, especially if the user's prior knowledge about the task is poor,
learning kernel methods require him only to specify a family of
kernels. The learning algorithm then selects both the specific kernel
out of that family, and the hypothesis defined with respect to that
kernel.

There is a large body of literature dealing with various aspects of
the problem of learning kernels, including theoretical questions,
optimization problems related to this problem, and experimental
results
\cite{lanckriet,micchelli_and_pontil,argyriou_colt,argyriou_icml,shai,ong,lewis_et_al,zienO07,jebara04,bach,l2reg,ying,nlk}.
Some of this previous work considers families of Gaussian kernels
\cite{micchelli_and_pontil} or hyperkernels \cite{ong}.  Non-linear
combinations of kernels have been recently considered by
\cite{varma,bach,nlk}.  But, the most common family of kernels
examined is that of non-negative combinations of some fixed kernels
constrained by a trace condition, which can be viewed as an $L_1$
regularization \cite{lanckriet}, or by an $L_2$ regularization
\cite{l2reg}.

This paper presents several novel generalization bounds for the
problem of learning kernels for the family of convex combinations of
base kernels or linear combinations with an $L_2$ constraint.  One of
the first learning bounds given by Lanckriet et al.\ \cite{lanckriet}
for the family of convex combinations of $p$ base kernels is similar
to that of Bousquet and Herrmann \cite{bousquet_and_herrmann} and has
the following form: $R(h) \leq \widehat R_\rho(h) +
O\big(\frac{1}{\sqrt{m}} \sqrt{\max_{k = 1}^p \text{Tr}(\K_k) \max_{i
    = 1}^p (\| \K_k \|/\Tr(\K_k))/\rho^2}\big)$ where $R(h)$ is the
generalization error of a hypothesis $h$, $R_\rho(h)$ is the fraction of
training points with margin less than or equal to $\rho$ and $\K_k$ is
the kernel matrix associated to the $k$th base kernel. This bound was
later shown by Srebro and Ben-David \cite{shai} to be always larger
than one. Another bound by Lanckriet et al.\ \cite{lanckriet} for the
family of linear combinations of base kernels was also shown by the
same authors to be always larger than one. 

But Lanckriet et al.\ \cite{lanckriet} also presented a multiplicative
bound for convex combinations of base kernels that is of the form
$R(h) \leq \widehat R_\rho(h) +
O\Big(\sqrt{\frac{p/\rho^2}{m}}\Big)$. This bound converges and can
perhaps be viewed as the first informative generalization bound for
this family of kernels. However, the dependence of the bound on the
number of kernels $p$ is multiplicative which therefore does not
encourage the use of too many base kernels.  Srebro and Ben-David
\cite{shai} presented a generalization bound based on the
pseudo-dimension of the family of kernels that significantly improved
on this bound.  Their bound has the form $R(h) \leq \widehat R_\rho(h)
+ \widetilde O\Big(\sqrt{\frac{p + R^2/\rho^2}{m}}\Big)$, where the
notation $\widetilde O(\cdot)$ hides logarithmic terms and where $R$
is an upper bound on $K_k(x, x)$ for all points $x$ and base kernels
$k_k$, $k \in [1, p]$.  Thus, disregarding logarithmic terms, their
bound is only additive in $p$.  Their analysis also applies to other
families of kernels. Ying and Campbell \cite{ying} also give
generalization bounds for learning kernels based on the notion of
Rademacher chaos complexity and the pseudo-dimension of the family of
kernels used. It is not clear however how their bound compares to that
of Srebro and Ben-David. We present new generalization bounds for the
family of convex combinations of base kernels that have only a
logarithmic dependency on $p$. Our learning bound is based on a
careful analysis of the Rademacher complexity of the hypothesis set
considered and has the form: $R(h) \leq \widehat R_\rho(h) +
O\Big(\sqrt{\frac{(\log p) R^2/\rho^2}{m}}\Big)$. Our bound is simpler
and contains no other extra logarithmic term. Thus, this represents a
substantial improvement over the previous best bounds for this
problem. Our bound is also valid for a very large number of kernels,
in particular for $p \gg m$, while the previous bounds were not
informative in that case.

We also present new generalization bounds for the family of linear
combinations of base kernels with an $L_2$ regularization.  We had
previously given a stability bound for an algorithm extending kernel
ridge regression to learning kernels that had an additive dependency
with respect to $p$ \cite{l2reg} assuming a technical condition of
orthogonality on the base kernels.  The complexity term of our bound
was of the form $O(1/\sqrt{m} + \sqrt{p/m})$. Our new learning bound
admits only a mild dependency of $p^{1/4}$ on the number of base kernels.

The next section (Section~\ref{sec:preliminaries}) defines the family
of kernels and hypothesis sets we examine. Section~\ref{sec:rademacher1}
presents a bound on the Rademacher complexity of the class of convex
combinations of base kernels with an  $L_1$ constraint and a generalization
bond for binary classification directly derived from that result. Similarly,
Section~\ref{sec:rademacher2} presents first a bound on the Rademacher
complexity, then a generalization bound for the case of an of $L_2$ 
regularization.

\section{Preliminaries}
\label{sec:preliminaries}

Most learning kernel algorithms are based on a hypothesis set derived from
convex combinations of a fixed set of kernels $K_1, \ldots, K_p$:
\begin{equation}
H_p = \Big\{\sum_{i = 1}^m \alpha_i K(x_i, \cdot)\colon K = \sum_{k = 1}^p \mu_k K_k, \mu_k \geq 0, \sum_{k = 1}^p \mu_k = 1, \Alpha^\top \K \Alpha \leq 1/\rho^2 \Big\}.
\end{equation}
Note that linear combinations with possibly negative mixture weights have
also been considered in the literature, e.g., \cite{lanckriet}, however
these combinations do not ensure that the combined kernel is PDS.

We also consider the hypothesis set $H'_p$ based on a $L_2$ condition
on the vector $\Mu$ and defined as follows:
\begin{equation}
H'_p = \Big\{\sum_{i = 1}^m \alpha_i K(x_i, \cdot)\colon K = \sum_{k = 1}^p \mu_k K_k, \mu_k \geq 0, \sum_{k = 1}^p \mu^2_k = 1, \Alpha^\top \K \Alpha \leq 1/\rho^2 \Big\}.
\end{equation}
We bound the empirical Rademacher complexity $\h \R_S(H_p)$ or $\h
\R_S(H'_p)$ of these families for an arbitrary sample $S$ of size $m$,
which immediately yields a generalization bound for learning kernels
based on this family of hypotheses. For a fixed sample $S = (x_1,
\ldots, x_m)$, the empirical Rademacher complexity of a hypothesis set
$H$ is defined as
\begin{equation}
  \h \R_S(H) = \frac{1}{m} \E_\sigma \Big[\sup_{h \in H} \sum_{i=1}^m \sigma_i h(x_i) \Big].
\end{equation}
The expectation is taken over $\sigma = (\sigma_1, \ldots, \sigma_n)$
where $\sigma_i$s are independent uniform random variables taking
values in $\set{-1, +1}$.

Let $h \in H_p$, then
\begin{align}
h(x) = \sum_{i = 1}^m \alpha_i K(x_i, x) = \sum_{k = 1}^p \sum_{i = 1}^m \mu_k \alpha_i K_k(x_i, x) = \w \cdot \P(x),
\end{align}
where $\w = 
\Bigg[ 
\begin{smallmatrix}
\w_1\\
\vdots\\
\w_p
\end{smallmatrix}\Bigg]$
with $\w_k = \mu_k \sum_{i = 1}^m \alpha_i \P_k(x_i)$ and
$\P(x) = 
\Bigg[ 
\begin{smallmatrix}
\P_1(x)\\
\vdots\\
\P_p(x)
\end{smallmatrix}\Bigg]$ with $\P_k = K_k(x, \cdot)$, for all $k \in
[1, p]$.

\section{Rademacher complexity bound for $H_p$}
\label{sec:rademacher1}

\begin{theorem}
\label{th:1}
For any sample $S$ of size $m$, the Rademacher complexity of the
hypothesis set $H_p$ can be bounded as follows:
\begin{equation}
\h \R_S(H_p) \leq \frac{\| \Tau \|_r}{m \rho}  \quad \text{with } \Tau = (\sqrt{r \Tr[\K_1]}, \ldots, \textstyle \sqrt{r \Tr[\K_p]})^\top,
\end{equation}
for any even integer $r > 0$. If additionally, $K_k(x, x) \leq R^2$
for all $x \in X$ and $k \in [1, p]$, then, for $p > 1$,
\begin{equation*}
\h \R_S(H_p) \leq \sqrt{\frac{2e \lceil \log p \rceil R^2/ \rho^2}{m}}.
\end{equation*}

\end{theorem}
\begin{proof}
Fix a sample $S$, then $\h \R_S(H_p)$ can be bounded as follows for
the hypothesis set of kernel learning algorithms for any $q, r > 1$ with
$1/q + 1/r = 1$:
\begin{align*}
\h \R_S(H_p) 
& = \frac{1}{m} \E_\sigma \Big[\sup_{h \in H_p} \sum_{i=1}^m \sigma_i h(x_i) \Big]\\
& = \frac{1}{m} \E_\sigma \Big[\sup_{\w} \w \cdot \sum_{i = 1}^m \sigma_i \P(x_i) \Big]\\
& \leq \frac{1}{m} \E_\sigma \Big[\sup_{\w} \Big(\sum_{k = 1}^p \|\w_k\|^q \Big)^{1/q}  \Big(\sum_{k = 1}^p \Big\|\sum_{i = 1}^m \sigma_i \P_k(x_i) \Big\|^r \Big)^{1/r} \Big] \quad (\text{Lemma~\ref{lemma:1}})\\
& = \frac{1}{m} \bigg[ \sup_{\w} \Big(\sum_{k = 1}^p \|\w_k\|^q \Big)^{1/q} \bigg] \E_\sigma \bigg[ \Big(\sum_{k = 1}^p \Big\|\sum_{i = 1}^m \sigma_i \P_k(x_i) \Big\|^r \Big)^{1/r} \bigg].
\end{align*}

We bound each of these two factors separately. 
The first term can be bounded as follows.
\begin{align*}
\Big(\sum_{k = 1}^p \|\w_k\|^q \Big)^{1/q} 
& \leq \sum_{k = 1}^p (\|\w_k\|^q)^{1/q} \quad (\text{sub-additivity of } x \mapsto x^{1/q}, (1/q) < 1)\\
%& = \sum_{k = 1}^p \|\w_k\|\\
& = \sum_{k = 1}^p \mu_k \| \sum_{i = 1}^m \alpha_i \P_k(x_i) \|\\
& \leq \sqrt{\sum_{k = 1}^p \mu_k \| \sum_{i = 1}^m \alpha_i \P_k(x_i) \|^2} \quad (\text{convexity})\\
& = \sqrt{\sum_{k = 1}^p \mu_k \Alpha^\top \K_k \Alpha} = \sqrt{\Alpha^\top \K \Alpha} \leq 1/\rho.
\end{align*}
We bound the second term as follows:
\begin{align*}
\E_\sigma \bigg[ \Big(\sum_{k = 1}^p \big\|\sum_{i = 1}^m \sigma_i \P_k(x_i) \big\|^r \Big)^{1/r} \bigg]
& \leq  \bigg( \E_\sigma \Big[ \sum_{k = 1}^p \big\|\sum_{i = 1}^m \sigma_i \P_k(x_i) \big\|^r \Big] \bigg)^{1/r}   \quad (\text{Jensen's inequality})\\
& =  \bigg( \sum_{k = 1}^p  \E_\sigma \Big[\big\|\sum_{i = 1}^m \sigma_i \P_k(x_i) \big\|^r \Big] \bigg)^{1/r} 
\end{align*}
Suppose that $r$ is an even integer, $r = 2r'$. Then, we can bound the expectation as follows:
\begin{align*}
\E_\sigma \Big[\big\|\sum_{i = 1}^m \sigma_i \P_k(x_i) \big\|^r \Big] 
& = \E_\sigma \Big[ \Big(\sum_{i, j = 1}^m \sigma_i \sigma_j K_k(x_i, x_j) \Big)^{r'} \Big]\\
% & = \sum_{\substack{1 \leq i_1, \ldots, i_{r'} \leq m\\1 \leq j_1, \ldots, j_{r'} \leq m}} \E_\sigma \Big[ \sigma_{i_1} \sigma_{j_1} \cdots \sigma_{i_{r'}} \sigma_{j_{r'}}\Big] K_k(x_{i_1}, x_{j_1}) \cdots K_k(x_{i_{r'}}, x_{j_{r'}}) \\
& \leq \sum_{\substack{1 \leq i_1, \ldots, i_{r'} \leq m\\1 \leq j_1, \ldots, j_{r'} \leq m}} \Big| \E_\sigma \Big[ \sigma_{i_1} \sigma_{j_1} \cdots \sigma_{i_{r'}} \sigma_{j_{r'}}\Big] \Big| \,  \Big| K_k(x_{i_1}, x_{j_1}) \cdots K_k(x_{i_{r'}}, x_{j_{r'}}) \Big| \\
& \leq \sum_{\substack{1 \leq i_1, \ldots, i_{r'} \leq m\\1 \leq j_1, \ldots, j_{r'} \leq m}}  \Big| \E_\sigma \Big[ \sigma_{i_1} \sigma_{j_1} \cdots \sigma_{i_{r'}} \sigma_{j_{r'}}\Big]  \Big| (K_k(x_{i_1}, x_{i_1}) \cdots K_k(x_{i_{r'}}, x_{i_{r'}}))^{1/2} \\
& \qquad \qquad \qquad \qquad (K_k(x_{j_1}, x_{j_1}) \cdots K_k(x_{j_{r'}}, x_{j_{r'}}))^{1/2} \quad (\text{Cauchy-Schwarz})\\
& = \sum_{s_1 + \ldots + s_m = 2r'}  \tbinom{2r'}{s_1, \ldots, s_m} \Big| \E_\sigma \Big[\sigma_1^{s_1} \cdots \sigma_m^{s_m}\Big]  \Big| K_k(x_1, x_1)^{s_1/2} \cdots K_k(x_m, x_m)^{s_m/2}.
\end{align*}
Since $\E[\sigma_{i}] = 0$ for all $i$ and since the Rademacher
variables are independent, we can write $\E[\sigma_{i_1} \ldots
\sigma_{i_l}] = \E[\sigma_{i_1}] \cdots \E[\sigma_{i_l}] = 0$ for any
$l$ distinct variables $\sigma_{i_1}, \ldots, \sigma_{i_l}$.  Thus,
 $\E_\sigma \Big[\sigma_1^{s_1}
\cdots \sigma_1^{s_m}\Big] = 0$ unless all $s_i$s are even, in which
case $\E_\sigma \Big[\sigma_1^{s_1} \cdots \sigma_m^{s_m}\Big] =
1$. Therefore, the following inequality holds:\footnote{We use the
  following rather rough inequality:
\begin{equation*}
\tbinom{2r'}{2 t_1, \ldots, 2 t_m} = \frac{(2r')!}{(2t_1)! \cdots (2t_m)!} 
\leq \frac{(2r')!}{(t_1)! \cdots (t_m)!} 
\leq \frac{(2r') \cdots (r' + 1) \cdot r'!}{(t_1)! \cdots (t_m)!}
\leq \frac{(2r')^{r'} \cdot r'!}{(t_1)! \cdots (t_m)!}
= (2r')^{r'} \tbinom{r'}{t_1, \ldots, t_m}.
\end{equation*}
}
\begin{align*}
\E_\sigma \Big[\big\|\sum_{i = 1}^m \P_k(x_i) \big\|^r \Big] 
& \leq \sum_{2t_1 + \ldots + 2t_m = 2r'}  \tbinom{2r'}{2 t_1, \ldots, 2 t_m} K_k(x_1, x_1)^{t_1} \cdots K_k(x_m, x_m)^{t_m}\\
& \leq (2r')^{r'} \sum_{t_1 + \ldots + t_m = r'}  \tbinom{r'}{t_1, \ldots, t_m} K_k(x_1, x_1)^{t_1} \cdots K_k(x_m, x_m)^{t_m}\\
& = (2r' \Tr[\K_k])^{r'} = (r \Tr[\K_k])^{r/2}.
\end{align*}
Thus, the Rademacher complexity is bounded by
\begin{equation}
\h \R_S(H_p) \leq \frac{\| \Tau \|_r}{m \rho}  \quad \text{with } \Tau = (\sqrt{r \Tr[\K_1]}, \ldots, \textstyle \sqrt{r \Tr[\K_p]})^\top,
\end{equation}
for any even integer $r$.  

Assume that $K_k(x, x) \leq R^2$ for all $x \in X$ and $k \in [1, p]$.
Then, $\Tr[\K_k] \leq m R^2$ for any $k \in [1, p]$, thus
the Rademacher complexity can be bounded as follows
\begin{equation*}
\h \R_S(H_p) \leq \frac{1}{m \rho} (p (\sqrt{r m R^2})^r)^{1/r} = p^{1/r} r^{1/2} \sqrt{\frac{R^2/ \rho^2}{m}}.
\end{equation*}
For $p > 1$, the function $r \mapsto p^{1/r} r^{1/2}$ reaches its
minimum at $r_0 = 2 \log p$. This gives
\begin{equation*}
\h \R_S(H_p) \leq \sqrt{\frac{2e \lceil \log p \rceil R^2/ \rho^2}{m}}.
\end{equation*}
\end{proof}

It is likely that the constants in the bound of theorem can be further
improved. We used a very rough upper bound for the multinomial
coefficients. A finer bound using Sterling's approximation should
provide a better result. Remarkably, the bound of the theorem has a
very mild dependency with respect to $p$.

The theorem can be used to derive generalization bounds for learning
kernels in classification, regression, and other tasks. We briefly
illustrate its application to binary classification where the labels
$y$ are in $\set{-1, +1}$.  Let $R(h)$ denote the generalization error
of $h \in H_p$, that is $R(h) = \Pr[y h(x) < 0]$. For a training
sample $S = ((x_1, y_1), \ldots, (x_m, y_m))$ and any
$\rho > 0$, let $\h R_\rho(h)$ denote the fraction of the training
points with margin less than or equal to $\rho$, that is $\h R_\rho(h)
= \tfrac{1}{m} \sum_{i = 1}^m 1_{y_i h(x_i) \leq \rho}$. Then, the 
following result holds.
\begin{corollary}
  For any $\delta > 0$, with probability at least $1 - \delta$, the
  following bound holds for any $h \in H_p$:
\begin{equation}
R(h) \leq \h R_\rho(h) + \frac{2 \| \Tau \|_r}{m \rho} + 2 \sqrt{\frac{\log\frac{2}{\delta}}{2m}}.
\end{equation}
with $\Tau = (\sqrt{r \Tr[\K_1]}, \ldots, \textstyle \sqrt{r \Tr[\K_p]})^\top$,
for any even integer $r > 0$. If additionally, $K_k(x, x) \leq R^2$
for all $x \in X$ and $k \in [1, p]$, then, for $p > 1$,
\begin{equation*}
R(h) \leq \h R_\rho(h) + 2 \sqrt{\frac{2e \lceil \log p \rceil R^2/ \rho^2}{m}} + 2 \sqrt{\frac{\log\frac{2}{\delta}}{2m}}.
\end{equation*}
\end{corollary}

\begin{proof}
  With our definition of the Rademacher complexity, for any $\delta >
  0$, with probability at least $1 - \delta$, the following
  bound holds for any $h \in H_p$ \cite{koltchinskii_and_panchenko,bartlett}:
\begin{equation}
R(h) \leq \h R_\rho(h) + 2 \h \R_S(H_p) + 2 \sqrt{\frac{\log\frac{2}{\delta}}{2m}}.
\end{equation}
Plugging in the bound on the empirical Rademacher complexity given
by Theorem~\ref{th:1} yields the statement of the corollary.
\end{proof}

The corollary gives a generalization bound for learning
kernels with $H_p$ that is in
\begin{equation}
O\bigg(\sqrt{\frac{(\log p) \ R^2/\rho^2}{m}}\bigg).
\end{equation}
In comparison, the bound for this problem given by Srebro and Ben-David
 \cite{shai} using the pseudo-dimension has a stronger
dependency with respect to $p$ and is more complex:
\begin{equation}
O\Bigg(\sqrt{8 
\frac{2 + p \log \frac{128 em^3 R^2}{\rho^2p} + 256 \frac{R^2}{\rho^2} \log \frac{\rho em}
{8R} \log \frac{128 m R^2}{\rho^2} } {m}}\Bigg).
\end{equation}
This bound is also not informative for $p > m$.

\section{Rademacher complexity bound for $H'_p$}
\label{sec:rademacher2}

\begin{theorem}
For any sample $S$ of size $m$, the Rademacher complexity of the
hypothesis set $H'_p$ can be bounded as follows:
\begin{equation}
\h \R_S(H'_p) \leq \frac{\| \Tau \|_r}{m \rho}  \quad \text{with } \Tau = (\sqrt{r \Tr[\K_1]}, \ldots, \textstyle \sqrt{r \Tr[\K_p]})^\top,
\end{equation}
for any even integer $0 < r \leq 4$. If additionally, $K_k(x, x) \leq R^2$
for all $x \in X$ and $k \in [1, p]$, then, for any $p \geq 1$,
\begin{equation*}
\h \R_S(H'_p) \leq 2 p^{1/4} \sqrt{\frac{R^2/ \rho^2}{m}}.
\end{equation*}
\end{theorem}
This bound also hold without the condition $\mu_k \geq 0 $, $k \in [1,
p]$, on the hypothesis set $H'_p$.

\begin{proof}
We can proceed as in the proof for bounding the Rademacher complexity
of $H_p$, except for bounding the following term:
\begin{align*}
\Big(\sum_{k = 1}^p \|\w_k\|^q \Big)^{1/q} 
& = \Big[ \sum_{k = 1}^p \mu_k^q (\Alpha^\top \K_k \Alpha)^{q/2} \Big]^{1/q}\\
& = \Big[ \sum_{k = 1}^p \mu_k^2 (\mu_k^{\frac{2 (q - 2)}{q}} \Alpha^\top \K_k \Alpha)^{q/2} \Big]^{1/q}\\
& \leq \Big[  (\sum_{k = 1}^p \mu_k^2 \, \mu_k^{\frac{2 (q - 2)}{q}} \Alpha^\top \K_k \Alpha)^{q/2} \Big]^{1/q} \quad (\text{convexity})\\
& = \sqrt{\sum_{k = 1}^p \mu_k^{\frac{4 (q - 1)}{q}} \Alpha^\top \K_k \Alpha}.
\end{align*}
Assume now that $q > 4/3$, which implies $\frac{4 (q - 1)}{q} < 1$.
Then, since $\mu_k \in [0, 1]$, this implies $\mu_k^{\frac{4 (q - 1)}{q}} \leq \mu_k$. Thus, for any $q > 4/3$, we can write:
\begin{equation*}
\Big(\sum_{k = 1}^p \|\w_k\|^q \Big)^{1/q} 
\leq \sqrt{\sum_{k = 1}^p \mu_k \Alpha^\top \K_k \Alpha}\\
= \sqrt{\Alpha^\top \K \Alpha} \leq 1/\rho^2.
\end{equation*}
Taking the limit $q \rightarrow 4/3$ shows that the inequality is also
verified for $q = 4/3$. Thus, as in the proof for $H_p$, the
Rademacher complexity can be bounded as follows
\begin{equation}
\h \R_S(H'_p) \leq \frac{\| \Tau \|_r}{m \rho}  \quad \text{with } \Tau = (\sqrt{r \Tr[\K_1]}, \ldots, \textstyle \sqrt{r \Tr[\K_p]})^\top,
\end{equation}
but here $r$ is an even integer such that $1/r = 1 - 1/q \geq 1 - 3/4
= 1/4$, that is $r \leq 4$. Assume that $K_k(x, x) \leq R^2$ for all
$x \in X$ and $k \in [1, p]$. Then, $\Tr[\K_k] \leq m R^2$ for any $k
\in [1, p]$, thus, for $r = 4$, the Rademacher complexity can be
bounded as follows
\begin{equation*}
\h \R_S(H'_p) \leq \frac{1}{m \rho} (p (\sqrt{4 m R^2})^{4})^{1/4} = 2 p^{1/4} \sqrt{\frac{R^2/ \rho^2}{m}}.
\end{equation*}
\end{proof}

Thus, in this case, the bound has a mild dependence
($\sqrt[4]{\cdot}$) on the number of kernels $p$. Proceeding as in the
$L_1$ case leads to the following margin bound in binary
classification.

\begin{corollary}
  For any $\delta > 0$, with probability at least $1 - \delta$, the
  following bound holds for any $h \in H_p$:
\begin{equation}
R(h) \leq \h R_\rho(h) + \frac{2 \| \Tau \|_r}{m \rho} + 2 \sqrt{\frac{\log\frac{2}{\delta}}{2m}}.
\end{equation}
with $\Tau = (\sqrt{r \Tr[\K_1]}, \ldots, \textstyle \sqrt{r \Tr[\K_p]})^\top$,
for any even integer $r \in \set{2, 4}$. If additionally, $K_k(x, x) \leq R^2$
for all $x \in X$ and $k \in [1, p]$, then, for any $p \geq 1$,
\begin{equation*}
R(h) \leq \h R_\rho(h) + 4 p^{1/4} \sqrt{\frac{R^2/ \rho^2}{m}} + 2 \sqrt{\frac{\log\frac{2}{\delta}}{2m}}.
\end{equation*}
\end{corollary}

\section{Conclusion}

We presented several new generalization bounds for the problem of
learning kernels with non-negative combinations of base kernels.  Our
bounds are simpler and significantly improve over previous
bounds. Their very mild dependency on the number of kernels seems to
suggest the use of a large number of kernels for this problem. Our
experiments with this problem in regression using a large number of
kernels seems to corroborate this idea \cite{l2reg}. Much needs to be
done however to combine these theoretical findings with the
somewhat disappointing performance observed in practice in most
learning kernel experiments \cite{cortes09}.

\bibliographystyle{plain}
\bibliography{lk}

\newpage
\appendix
\section{Lemma \ref{lemma:1}}
The following lemma is a straightforward version of 
H\"older's inequality.

\begin{lemma}
\label{lemma:1}
Let $q, r > 1$ with $1/q + 1/r = 1$. Then, the following result
similar to H\"older's inequality holds:
\begin{equation}
| \w \cdot \P(x) | \leq \Big(\sum_{k = 1}^p \|\w_k\|^q \Big)^{1/q}  \Big(\sum_{k = 1}^p \Big\| \P(x) \Big\|^r \Big)^{1/r}.
\end{equation}
\end{lemma}
\begin{proof}
Let $\Psi_q(\w) = (\sum_{k = 1}^p \| \w_k \|^q)^{1/q}$ and
$\Psi_r(\P(x)) = (\sum_{k = 1}^p \| \P_k(x) \|^r)^{1/r}$, then
\begin{align*}
\frac{|\w \cdot \P(x)|}{\Psi_q(\w) \Psi_r(\P(x))} 
& = \Big| \sum_{k = 1}^p \frac{\w_k}{\Psi_q(\w)} \cdot  \frac{\P_k(x)}{\Psi_r(\P(x))}\Big|\\
& \leq \sum_{k = 1}^p \Big| \frac{\w_k}{\Psi_q(\w)} \cdot  \frac{\P_k(x)}{\Psi_r(\P(x))}\Big|\\
& \leq \sum_{k = 1}^p \frac{\| \w_k \| }{\Psi_q(\w)} \cdot  \frac{\| \P_k(x) \|}{\Psi_r(\P(x))} & (\text{Cauchy-Schwarz})\\
& \leq \sum_{k = 1}^p \frac{1}{q}\frac{\| \w_k \|^q }{\Psi_q(\w)^q} + \frac{1}{r}  \frac{\| \P_k(x) \|^r}{\Psi_r(\P(x))^r} & (\text{Young's inequality})\\
& = \frac{1}{q} + \frac{1}{r} = 1.
\end{align*}
\end{proof}

\end{document}